\documentclass[runningheads]{llncs}
\usepackage[T1]{fontenc}
\usepackage{graphicx}
\usepackage{booktabs}
\usepackage{tabularx}
\usepackage{subfig}
\usepackage[frozencache=true,cachedir=minted-ijclr_2024]{minted}
\setminted[prolog]{frame=lines,fontsize=\small,framesep=4mm,numbers=left,numberblanklines=false,numbersep=5pt,baselinestretch=1.2}
\setminted[prolog]{frame=lines,fontsize=\small,framesep=1mm,numbers=left,numberblanklines=false,numbersep=5pt,baselinestretch=1.2}
\begin{document}
\title{From model-based learning to model-free behaviour with Meta-Interpretive
Learning}
\titlerunning{From model-based learning to model-free behaviour}
%
\author{Stassa Patsantzis}
\authorrunning{S. Patsantzis}
%
\institute{University of Surrey, UK \email{s.patsantzis-@surrey.ac.uk} }
\maketitle              

\begin{abstract}

A "model" is a theory that describes the state of an environment and the effects
of an agent's decisions on the environment. A model-based agent can use its
model to predict the effects of its future actions and so plan ahead, but must
know the state of the environment. A model-free agent cannot plan, but can act
without a model and without completely observing the environment. An autonomous
agent capable of acting independently in novel environments must combine both
sets of capabilities. We show how to create such an agent with Meta-Interpretive
Learning used to learn a model-based Solver used to train a model-free
Controller that can solve the same planning problems as the Solver. We
demonstrate the equivalence in problem-solving ability of the two agents on grid
navigation problems in two kinds of environment: randomly generated mazes, and
lake maps with wide open areas. We find that all navigation problems solved by
the Solver are also solved by the Controller, indicating the two are equivalent.

\end{abstract}
\section{Introduction}

\begin{figure}[t]
	\centering
	\begin{tabular}{ll}
		\subfloat[Model-based Solver \label{fig:solver_cave}]{\includegraphics[width=0.29\columnwidth]{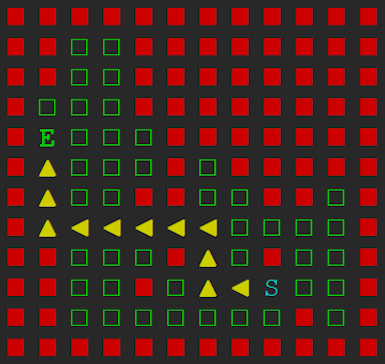}} &
		\quad
		\subfloat[Model-free Controller \label{fig:controller_cave}]{\includegraphics[width=0.29\columnwidth]{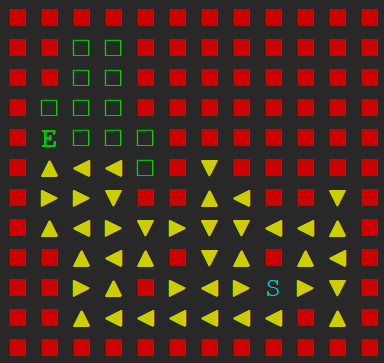}} \\
	\end{tabular}
	\caption{A model-based Solver can predict the effect of its future
	actions and plan ahead, but requires knowledge of its environment. A
	model-free Controller cannot plan and must explore its environment but
	can act without knowledge of the environment. Green tiles: passable; red
	tiles: unpassable; S: start; E: goal.}	\label{fig:solver_vs_controller}
\end{figure}

A \emph{model} is a theory that describes the possible states of an environment
and the way in which the state of the environment changes as a result of an
agent's actions. An \emph{environment} may be an abstract domain such as the set
of all lists or the Cartesian plane, a virtual environment, such as a computer
simulation or game, or a physical environment such as a real-world location. An
\emph{action} is a predicate that relates a description of the current state of
an environment to a new state; in other words, a transition relation over
states. An \emph{agent} is a program that generates a sequence of actions
connecting an initial state to a desired goal state; what we call a \emph{plan}.
To generate such a sequence of actions is to solve a \emph{planning problem}.  A
\emph{model-based} agent is an agent that solves planning problems given a
model. A \emph{model-free} agent is an agent that selects actions without a
model and, therefore, without a plan. We call a model-based agent a "solver for
planning problems" or, simply, \emph{solver} and a model-free agent a
\emph{controller}.

Figure \ref{fig:solver_vs_controller} illustrates the difference between
planning by a model-based Solver and acting by a model-free Controller. In the
Figure the environment is a map of a lake with islands (alternatively a cave
system) where red tiles represent unpassable locations, green tiles passable
locations and the letters $S$ and $E$ a starting and goal location respectively.
Yellow arrows show agent trajectories.

Planning with a model implies some knowledge of the structure of an environment.
The Solver in Figure \ref{fig:solver_cave} is initially given a map of the
environment, which it can fully observe, and uses this map to generate the set
of all possible moves between adjacent, passable map tiles. The Solver then uses
this navigation model to plan a direct path from $S$ to $E$. The Controller in
Figure \ref{fig:controller_cave} is not given the map of the environment, cannot
directly observe $S$ or $E$, and can only observe whether map tiles adjacent to
its current location are passable or unpassable. The Controller explores its
environment to find a path from $S$ to $E$.

A model-based solver and a model-free controller have complementary advantages
and disadvantages. An autonomous agent capable of acting independently in
arbitrary environments must combine the capabilities of both. In this paper we
show how the two sets of abilities can be combined by using Meta-Interpretive
Learning (MIL) \cite{Muggleton2014} to learn a solver, and use the solver to
generate examples to learn a controller that solves the same problems as the
solver.

Our ultimate goal is the development of an action-selection component of an
Autonomous Agent Architecture that must guide a mobile robot to survery missions
in environments with dynamic and unobserved features so that the autonomous
agent guiding the robot must combine the capabilities of both solver and
controller. Full details are reserved for future work.

\subsubsection{Inductive Logic Programming as model-based learning}
\label{Inductive Logic Programming as model-based learning}

\begin{figure}[t]
	\centering
	\subfloat[Grid navigation solver\label{tab:grid_solver}]{
		\begin{tabular}{l}
			$S(s_1,s_2) \leftarrow Step\_down(s_1,s_2).$ \\
			$S(s_1,s_2) \leftarrow Step\_left(s_1,s_2).$ \\
			$S(s_1,s_2) \leftarrow Step\_right(s_1,s_2).$ \\
			$S(s_1,s_2) \leftarrow Step\_up(s_1,s_2).$ \\
			$S(s_1,s_2) \leftarrow Step\_down(s_1,s_3),S(s_3,s_2).$ \\
			$S(s_1,s_2) \leftarrow Step\_left(s_1,s_3),S(s_3,s_2).$ \\
			$S(s_1,s_2) \leftarrow Step\_right(s_1,s_3),S(s_3,s_2).$ \\
			$S(s_1,s_2) \leftarrow Step\_up(s_1,s_3),S(s_3,s_2).$ 
		\end{tabular}
	}
	\subfloat[Grid navigation solver model\label{tab:grid_solver_model}]{
		\begin{tabular}{l}
			\\
			\\
			\\
			\\
			$Step\_down([id,x/y,t],[id,x/y^{-1},t')$ \\
			$Step\_left([id,x/y,t],[id,x^{-1}/y,t')$ \\
			$Step\_right([id,x/y,t],[id,x^{+1}/y,t'])$ \\
			$Step\_up([id,x/y,t],[id,x/y^{+1},t'])$ 
		\end{tabular}
	}\qquad
	\subfloat[Solver model instantiated to Maze A\label{tab:solver_actions_maze_a}]{
		\begin{tabular}{l}
			$Step\_right([maze\_a,0/6,s],[maze\_a,1/6,f])$ \\
			$Step\_down([maze\_a,2/6,f],[maze\_a,2/5,f])$ \\
			$Step\_left([maze\_a,2/0,f],[maze\_a,1/0,f])$ \\
			$Step\_left([maze\_a,1/0,f],[maze\_a,0/0,e])$ \\
			\% ... 56 more actions
		\end{tabular}
	}
	\begin{tabular}{r}
		\subfloat[Path in Maze A\label{fig:maze_a_solver}]{\includegraphics[width=0.22\columnwidth]{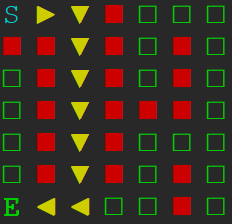}} \\
	\end{tabular}
	\subfloat[Solver model instantiated to Maze B\label{tab:solver_actions_maze_b}]{
		\begin{tabular}{l}
			$Step\_right([maze\_b,0/6,s],[maze\_b,1/6,f])$ \\
			$Step\_down([maze\_b,2/6,f],[maze\_b,2/5,f])$ \\
			$Step\_up([maze\_b,4/0,f],[maze\_b,4/1,f])$ \\
			$Step\_down([maze\_b,6/1,f],[maze\_b,6/0,e])$ \\
			\% ... 56 more actions
		\end{tabular}
	}
	\begin{tabular}{r}
		\subfloat[Path in Maze B\label{fig:maze_b_solver}]{\includegraphics[width=0.22\columnwidth]{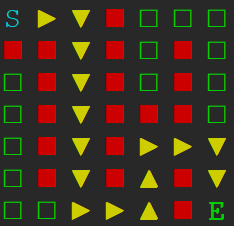}} \\
	\end{tabular}
	\caption{A Solver for grid navigation problems learned with MIL.} \label{fig:grid_solver}
\end{figure}

\emph{Plannning is the model-based approach to autonomous behaviour}
\cite{GeffnerAndBonet} in the sense that a planning agent's behaviour is derived
from a model and a planning problem by means of an inference procedure
(typically a search algorithm such as $A^*$). Inductive Logic Programming
(ILP)\cite{Muggleton1991} can be seen as the model-based approach to
\emph{machine learning} where the "model" is a background theory and a new
hypothesis is derived by an inference procedure given a set of examples. To
learn planning programs, i.e. solvers, with ILP, we can structure the background
theory as a set of action predicates and give a set of planning problems as
examples. For a solver to be general, it must be a recursive program, therefore
we use MIL, a form of ILP capable of learning recursion.

Figure \ref{fig:grid_solver} lists a grid navigation solver as a set of definite
clauses. The solver's body consists of eight clauses of the predicate $S/2$,
where each argument $s_i$ is a Prolog list representing an initial or end state
of the environment. Each clause of $S/2$ changes the environment state by moving
an agent one step to each of the four directions up, right, down or left,
recursively. The solver's model consists of the step actions $Step\_up,
Step\_right, Step\_down, Step\_left$, implemented as dyadic predicates, listed
in un-instantiated form in Figure \ref{tab:grid_solver_model}. The two arguments
of each action, shared with $S/2$, represent the state of the environment before
and after the action is taken as a list of first-order terms: $id$, a map
identifier, $x/y$, the coordinates of an agent on the map, and, $t$, the type of
map \emph{tile} at $x/y$, one of the constants $f,w,e,s$ for floor, wall, start
and end tile, respectively; wall tiles are not passable and so they are not
featured in actions' state arguments.

The solver's model must be instantiated to the coordinates and tile types of a
map before it can be used by the solver to solve that map. Two examples of
instantiated models are listed in Figures \ref{tab:solver_actions_maze_a},
\ref{tab:solver_actions_maze_b} for two mazes, Maze A and B. The two mazes are
identical save for the position of the end tile and so their instantiated models
are almost identical. Figures \ref{fig:maze_a_solver}, \ref{fig:maze_b_solver}
illustrate the solutions of the two mazes by the solver where each yellow arrow
corresponds to a step action in the indicated direction.

The solver in Figure \ref{fig:grid_solver} was learned by the MIL system Louise
\cite{Louise} as described in Section \ref{Learning a Model-Based Solver with
MIL}.

\subsubsection{Finite State Controllers for model-free behaviour}
\label{Finite State Controllers for model-free behaviour}

\begin{figure}[t]
	\centering
	\subfloat[Finite State Controller labels\label{tab:controller_model}]{
		\begin{tabular}{l}
			4 \textbf{Controller state labels:} $Q = \{q_0,q_1,q_2,q_3\}$ \\
			\textbf{4 Action labels:} $A = \{up,right,down,left\}$ \\
			\textbf{15 Observation labels:} $O = \{pppp,pppu,ppup,ppuu,pupp,pupu,puup,puuu,$ \\
			$uppp,uppu,upup,upuu,uupp,uupu,uuup\}$ \\
			\textbf{960 4-tuples:} $|Q \times O \times A \times Q|$	
		\end{tabular}
	} \qquad \qquad 
	\subfloat[FSC for Maze A\label{tab:maze_a_controller}]{
		\begin{tabular}{l}
			$(q_0,upuu,right,q_1) $ \\
			$(q_1,upup,right,q_1) $ \\
			$(q_1,uupp,down,q_2) $ \\
			$(q_2,pupu,down,q_2) $ \\
			$(q_2,ppup,left,q_3) $ \\
			$(q_3,upup,left,q_3) $ 
		\end{tabular} 
	}
	\begin{tabular}{r}
		\\
		\subfloat[Path in Maze A\label{fig:maze_a}]{\includegraphics[width=0.22\columnwidth]{tessera_2.png}} 
	\end{tabular}
	\subfloat[FSC for Maze B\label{tab:maze_b_controller}]{
		\begin{tabular}{l}
			$(q_1,upup,right,q_1).$ \\
			$(q_1,pupp,down,q_2).$ \\
			$(q_1,uupp,down,q_2).$ \\
			$(q_2,ppup,right,q_1).$ \\
			$(q_2,pupu,down,q_2).$ \\
			$(q_0,uppu,right,q_1).$ \\
			$(q_0,upuu,right,q_1).$ \\
			$(q_0,pupu,up,q_0).$ \\
			$(q_1,puup,up,q_0).$ 
		\end{tabular} 
	}
	\begin{tabular}{r}
		\\
		\subfloat[Path in Maze B\label{fig:maze_b}]{\includegraphics[width=0.215\columnwidth]{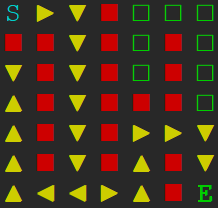}} \\
	\end{tabular}
	\caption{Finite State Controllers for mazes A, B trained with the Solver
	in Fig \ref{tab:grid_solver}.}
	\label{fig:controller_examples}
\end{figure}

When its model cannot be instantiated a model-based solver cannot plan. An
alternative is found in Finite State Controllers (FSCs) \cite{Bonet2009Short} a
formalisation of the concept of Finite State Machines from video games and
robotics. A FSC is a set of 4-tuples $(q,o,a,q')$ mapping pairs $(q,o)$ of
current \emph{controller} state\footnote{Controller states are not
\emph{environment} states.} and observation labels, to pairs $(a,q')$ of action
and next controller state labels. Figure \ref{fig:controller_examples} lists two
maze-solving FSCs for mazes A and B in Figure \ref{fig:grid_solver}, and their
sets of labels. Each observation label is a string $\{U,R,D,L\} \in \{u,p\}^2$
denoting whether map tiles in the directions up, right, down or left,
respectively, from the agent's current map location, are passable ($p$) or
unpassable ($u$). For example, the string $upuu$ denotes that only the tile to
the right of the agent is passable. Action labels denote that the agent must
move up, right, down or left. The two FSCs in Figure
\ref{fig:controller_examples} were learned by Louise from solutions of mazes A
and B generated by the solver in Figure \ref{tab:grid_solver}.

FSCs are model-free in that they have no representation of the environment
state, or the way actions modify that state; they are only a mapping between
pairs of labels. In particular, FSC observation labels represent an agent's
\emph{belief} about the state of the environment. Additionally, FSCs are
\emph{efficient} in that they do not need to search a large state-space like
planners, or, indeed, solvers. Conversely, an FSC is equivalent to a
Deterministic Finite State Transducer \cite{JurafskyMartin2000Short} a.k.a. a
Mealy Machine \cite{Mealy1955}, i.e. a Regular Automaton and so can only
generate one sequence of actions, e.g. the maze FSC in Figure
\ref{tab:maze_a_controller} can only solve mazes where the exit can be reached
by moving right, down, or left, like Maze A in Figure \ref{fig:maze_a_solver},
whereas the FSC in figure \ref{tab:maze_b_controller} can only solve mazes where
the exit can be reached by moving right, down or up, like Maze B in Figure
\ref{fig:maze_b_solver}.  Finally, the literature on FSCs does not describe a
concrete method to \emph{execute} FSCs, only their notation as sets of 4-tuples.

\subsubsection{Contributions}
\label{Contributions}

We identify FSCs as a promising framework for the development of model-free
controllers solving the same problems as model-based solvers in partially
observable environments. We make the following contributions.

\begin{itemize}
	\item We formalise a notation for model-based learning of planning
		problem solvers as a higher-order background theory for MIL.
	\item We extend the framework of FSCs to Nondeterministic FSCs.
	\item We implement a set of FSC \emph{executors}, stack-machines that
		take as input a set of FSC tuples and execute them in order.
	\item We implement a novel approach to Simultaneous Localisation and
		Mapping (SLAM) used by executors to avoid cycles.
	\item We show that a solver and a controller that solve the same
		problems can be learned by MIL using our model notation.
	\item We demonstrate empirically that our learned solver and controller
		can solve the same planning problems.
	\item We implement two new libraries in Prolog: \emph{Controller Freak},
		to learn controllers from solvers; and \emph{Grid Master} to
		solve navigation problems on grids.
\end{itemize}

Implementation code and experiment data are available at the following URL:
\url{https://github.com/stassa/ijclr24}.

\section{Related Work}
\label{Related Work}

\subsection{Meta-Interpretive Learning}
\label{Meta-Interpretive Learning}

Meta-Interpretive Learning (MIL)\cite{Muggleton2014,Muggleton2015} is a new form
of Inductive Logic Programming (ILP)\cite{Muggleton1991,Muggleton1994} where the
background theory is a higher-order program that includes both first- and
second-order definite clauses. Hypotheses learned by MIL are first-order
definite programs, i.e. logic programs in Prolog. Hypotheses are learned by SLD
Refutation of training examples, which can be both positive or negative, given
as Horn goals to a meta-interpreter capable of second-order SLD Resolution.
During Resolution, substitutions of second-order variables in the higher-order
background theory are derived which applied to their corresponding second-order
clauses yield the first-order clauses of a hypothesis. MIL has been shown
capable of learning recursion and of predicate invention without the limitations
of earlier ILP systems\cite{Patsantzis2021b}.

\subsection{Finite State Controllers}
\label{Finite State Controllers}

A Finite State Controller (FSC)\cite{Bonet2009Short} is a set of 4-tuples
$(q,o,a,q')$, where $q,q' \in Q, o \in O, a \in A$ are constants serving as
labels for controller states, observations and actions, respectively.  A tuple
$(q,o,a,q')$ corresponds to a mapping between pairs $(q,o)$ and $(a,q')$ and the
set of all tuples in an FSC to the \emph{transition function} $Q \times O
\rightarrow A \times Q$ that controls the actions of an agent, so that in
controller state $q$, when the observation label $o$ is input, the FSC outputs
action label $a$ and transitions to controller state $q'$. In
\cite{Bonet2009Short} an approach is described to derive FSCs from contingent
planning problems through a series of transformations to classical planning, via
conformant planning, problems \cite{GeffnerAndBonet}. The approach is
demonstrated in a number of domains including navigation on grids that we also
demonstrate here, but transformation betwen planning problems is subject to high
computational costs.

Compared to contingent plans and policies derived from Markov Decision Processes
(MDPs) and Partially Observable MDPs (POMDPS)\cite{Bonet2009Short} claim that
FSCs are: a) general, in that they can solve multiple problems in a domain; b)
compact, in that they only need comprise a few controller states; and c) robust
to uncertainty in initial and goal conditions and the effects of actions.
Generality of FSCs is formalised in \cite{Bonet2015Short} and their compactness
is self-evident. We could not find evidence of FSC robustness to uncertainty in
the literature; we leave the matter to future work.

To be used as an action selection mechanism for an autonomous agent, an FSC must
be executed in a loop by an external process. Such a process must connect the
FSC to an environment by passing observation and action labels between the FSC
and the environment. Such a mechanism has not, so far, been described in the
literature. In Section \ref{Implementation} we implement a set of
\emph{executors} that perform this function in conjunction with a virtual
environment for grid navigation problems. 

FSCs are equivalent to Deterministic Finite State Transducers (FSTs), a.k.a.
Mealy machines\footnote{This equivalence is not noted in earlier work.}. In an
FSC's tuples, observation and action labels are the inputs and outputs,
respectively, of a deterministic FST's transition function, and each pair $Q
\times O$ is mapped to exactly one pair $A \times Q$, such that there exists a
single tuple $(q,o,a,q')$ for each pair $(q,o)$ in an FSC. Being equivalent to
\emph{Deterministic} FSTs, in other words, Regular Automata, FSCs cannot deal
with ambiguities in the sense that they can only successfully solve planning
problems where the same action must always be taken in the same circumstances.
While this may be sufficient, or even desired, that is not always the case.

\begin{figure}[t]
	\centering
	\begin{tabular}{cc}
		\subfloat[Maze A \label{fig:maze_a_ambi}]{\includegraphics[width=0.20\columnwidth]{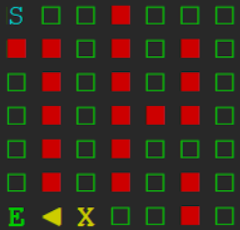}} &
		\subfloat[Maze B \label{fig:maze_b_ambi}]{\includegraphics[width=0.20\columnwidth]{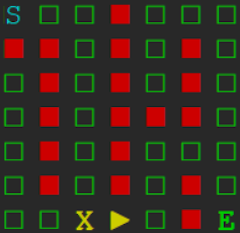}} \\
	\end{tabular}
	\caption{An example of ambiguity in a maze environment. The two mazes,
	Maze A and Maze B have identical structure but the end tile $E$ is
	placed in a different location. In Maze A, an agent must turn left at
	the yellow $X$ to reach the exit at $E$, but in Maze B it must turn
	right instead, as indicated by the yellow arrows.}.
	\label{fig:maze_ambiguity}
\end{figure}

As an example of the difficulty in dealing with ambiguities, consider the two
mazes A and B in Figure \ref{fig:maze_ambiguity}, identical in structure, but
with an end tile ($E$) placed in a different location. An FSC solving maze A
must turn left at the cell marked with a yellow $X$ in order to reach the exit,
therefore the same FSC will not be able to solve maze B where it must turn
\emph{right} at the $X$ to reach the exit\footnote{The same limitation obtains
when learning optimal policies from (PO)MDPs, as e.g. with Reinforcement
Learning where this limitation was recently identified as \emph{implicit partial
observability}\cite{Ranzato2021Short}, the characteristic of a domain where an
arbitrary instance of a problem does not include every observation possible in
all problem instances.}. In Section \ref{Framework} we extend FSCs to
\emph{Nondeterministic} FSCs and in Section \ref {Implementation} implement
executors with a stack that permit an FSC to backtrack or reverse course to
fully explore an environment and solve ambiguous environments.

\section{Framework}
\label{Framework}

\subsection{MIL Planning Model}
\label{MIL Planning Model}

To learn a solver for planning problems with MIL we structure the higher-order
background theory given to a MIL system as a model of the planning domain and
give as an example a planning problem. We call such a background theory and
example a \emph{MIL Planning Model}.

\begin{definition} (MIL Planning Model) A MIL Planning Model $\mathcal{T}$ is a
	5-tuple $\{\mathcal{F,S,A,M},e\}$ where:
	\begin{itemize} 
		\item $\mathcal{F}$ is a set of first-order terms that we call
			\emph{fluents}.
		\item $\mathcal{S}$ is a set of states $\{s_1, ..., s_m\}$,
			where each $s_i$ is a set of \emph{instances} of the
			fluents in $\mathcal{F}$, given as a Prolog list.
		\item $\mathcal{A}$ is a set of dyadic definite clauses with
			heads $A_1(s_{11},s_{12}), ..., A_n(s_{n1}, s_{n2})$,
			where each $s_{i1}, s_{i2} \in \mathcal{S}$, and
			$\mathcal{A}$ defines a \emph{transition relation}
			between states.
		\item $\mathcal{M}$ is the set containing the two second-order
			definite clauses: $P(x,y) \leftarrow Q(x,y)$ (called
			"Identity") and $P(x,y) \leftarrow Q(x,z), P(z,y)$
			("Tailrec") with $P,Q$ existentially quantified over the
			set of predicate symbols in $\mathcal{A}$ and the symbol
			$S/2$, and $x,y,z$ universally or existentially
			quantified over $\mathcal{S}$.
		\item $e$ is a Horn goal $\leftarrow S(s_{in},s_g)$ given as a
			training example, where $s_{in},s_g \in \mathcal{S}$ are
			an initial state and a goal state of a planning problem,
			respectively.
	\end{itemize} 
	We call $s_{i1},s_{i2}$ in an action $A_i$ the \emph{input} and
	\emph{output} states of $A_i$, respectively.
	\label{def:mil_planning_model}
\end{definition}

Clauses of action predicates in $\mathcal{A}$ are selected for Resolution by
unification so the fluents in actions' state-lists act as both preconditions and
effects, unlike in STRIPs-like planning models where there is a clear
distinction between the two.

Given a MIL Planning Model, a MIL system must return a \emph{planning
hypothesis} $H$ such that $\mathcal{A} \cup H \models e$. We say that $H$ solves
the planning problem in $e$. We call $H$ a \emph{solver} for $e$. In particular,
$H$ is a \emph{model-based} solver and its model is the set of ground instances
of clauses in $\mathcal{A}$\footnote{In this case the meaning of "model"
converges to that in First Order Logic.}.

\begin{definition} (MIL Planning Hypothesis) A MIL Planning Hypothesis $H$ is a
	set of first-order definite clause instances of \emph{Identity} and
	\emph{Tailrec} such that each $P$ in a head literal $P(x,y)$ in
	\emph{Identity} or \emph{Tailrec} and the last body literal $P(z,y)$ in
	\emph{Tailrec} is a literal of the predicate $S/2$ as in $e \in
	\mathcal{T}$, and each body literal $Q(x,y)$ or $Q(x,z)$ in
	\emph{Identity} and \emph{Tailrec} respectively is a literal of an
	action predicate $A_i(s_{i1},s_{i2})$ in $ \mathcal{A} \in \mathcal{T}$.
	If $\mathcal{A} \cup H \models e$, $H$ is a \emph{solver} for $e$.
	\label{def:mil_planning_hypothesis}
\end{definition}

\subsection{Nondeterministic Finite State Controllers}
\label{Nondeterministic Finite State Controllers}

We extend the definition of Finite State Controllers, adapted, in Definition
\ref{def:det_fscs}, from \cite{Bonet2009Short}, to \emph{Nondeterministic}
Finite State Controllers.

\begin{definition} (Finite State Controller) Let $Q$ be a set of controller
	state labels, $O$ a set of observation labels, $A$ a set of action
	labels, and let $C$ be a function $Q \times O \rightarrow A \times Q$.
	$C$ is a Finite State Controller, denoted as the set of 4-tuples
	$(q,o,a,q')$ where $q, q' \in Q, o \in O, a \in A$ and $(q,o)
	\rightarrow (a,q') \in C$. Note that all FSC labels are constants, or
	strings of constants.
	\label{def:det_fscs}
\end{definition}

\begin{definition} (Nondeterministic Finite State Controller) Let $Q,O,A$ be  as
	in Definition \ref{def:det_fscs}, let $R = Q \times O \times A \times
	Q$, and let $C \subseteq R$. We call $C$ a Nondeterministic Finite State
	Controller for $R$. We denote $C$ as the set of 4-tuples $(q,o,a,q')$
	that correspond to the subset of $R$ in $C$, where $q, q' \in Q, o \in
	O, a \in A$.
	\label{def:nondet_fscs}
\end{definition}

We will henceforth refer to Nondeterministic FSCs simply as FSCs, only
clarifying the determinism property where it is not clear from the context. We
call a sequence of FSC tuples the FSC's \emph{behaviour}:

\begin{definition} (FSC Behaviour) Let $C$ be a Nondeterministic FSC and
	$B = \langle T_1, ..., T_n \rangle$ a sequence of 4-tuples in $C$, such
	that for every $t_i, t_{i+1} \in B$ if $t_i = (q,o,a,q')$ then $t_{i+i}
	= (q',o',a',q'') $. We call $B$ the \emph{behaviour} of (an agent
	controlled by) $C$.
	\label{def:fsc_behaviour}
\end{definition}

\section{Implementation}
\label{Implementation}

\subsection{Controller Freak}
\label{Controller Freak}

\begin{table}[t]
	\centering
	\begin{tabular}{l}
		$Step\_down([id,x/y,t,[q|Qs],[o|Os],[a|As],[q'|Qs']],[id,x/y^{-1},t',Qs,Os,As,Qs'])$ \\
		$Step\_left([id,x/y,t,[q|Qs],[o|Os],[a|As],[q'|Qs']],[id,x^{-1}/y,t',Qs,Os,As,Qs'])$ \\
		$Step\_right([id,x/y,t,[q|Qs],[o|Os],[a|As],[q'|Qs']],[id,x^{+1}/y,t',Qs,Os,As,Qs'])$ \\
		$Step\_up([id,x/y,t,[q|Qs],[o|Os],[a|As],[q'|Qs']],[id,x/y^{+1},t',Qs,Os,As,Qs'])$ \\
	\end{tabular}
	\caption{Grid navigation solver model with controller labels
	used to generate FSC behaviours to learn a grid navigation FSC with
	MIL.}
	\label{tab:solver_fsc_model}
\end{table}

We implement, in Prolog, a new library called \emph{Controller Freak} that
defines Prolog predicates to learn FSCs from solvers, by MIL. Predicates in
Controller Freak take as input a set of planning problems $E$ and a solver $H$
where the set of fluents $F$ in actions $\mathcal{A}$ in $H$'s model includes
the sets of labels $Q, O, A$ that must be accepted by $C$, the FSC to be
learned. A set of actions $\mathcal{A}$ extending the grid navigation solver
model in Figure \ref{tab:grid_solver_model} with FSC labels is listed in Table
\ref{tab:solver_fsc_model}.

Solving each $e \in E$ with $H$ generates sequences of labels corresponding to
example behaviours $B$ of $C$ used to compose the following MIL-learning
problem:

\begin{itemize}
	\item First-order Background theory: $Q \times O \times A \times Q$
	\item Second-Order Background theory: $\{Identity, Tailrec\}$
	\item Examples: Set of FSC behaviours $B$ 
\end{itemize}

Solving this MIL-learning problem results in a set of definite clauses, $C_p$,
that represent $C$ in the form of a recursive logic program. Further processing
by Controller Freak predicates transforms $C_p$ into the set of 4-tuples of $C$.

\subsection{Grid Master}
\label{Grid Master}

We implement, in Prolog, a library for grid navigation problems called
\emph{Grid Master}, defining predicates in three modules: a) A \emph{Map} module
is used to manage grid-style maps in text format, where each "cell" of a map is
a string. A generator for maze maps called \emph{MaGe} is included with Grid
Master. b) An \emph{Action Generator} module is used to generate actions for a
MIL Planning Model given a map. c) A \emph{Map Display} module is used to
display maps and the progress of a grid navigation agent through a map, as in
the map figures in this paper.

Grid Master allows the definition of game-like virtual environments with which
an agent can interact to solve a map. An initial \emph{Basic Environment} is
included that manages navigation on mazes, and similar grids with a start and
end location. The Basic Environment a) initialises an agent's starting location
on a map, b) receives action labels from, and returns observation labels to, a
controller (via an executor) as the agent navigates the map, and c) determines
whether the agent has reached the end location on the map. Additionally, the
Basic Environment can receive as input a sequence of action labels from a
solver, and "play them back" to determine whether the solver has reached the end
location.

\subsection{Executors}
\label{Executors}

An \emph{executor} is an interpreter for FSCs implemented as a loop that passes
a current pair $(q,o)$ to an FSC, receives a pair $(a,q')$ and passes the action
label $a$ of this pair to an environment to obtain a new observation label $o$,
until a goal state is reached. Initial and goal states are represented as
fluents not directly observed by the executor, or the FSC it is executing.
Instead the environment, e.g. the Basic Environment, or the sensor and actor
models on a robot, translate current environment states to observation labels
and action labels to new environment states.

We define two kinds of executors: backtracking and reversing; and two variants,
with and without SLAM.

\begin{figure}[t]
	\begin{minted}{prolog}
	executor_bck(Gs,_Q0,_O,As,Gs,As):- !.
	executor_bck(Fs,Q0,O,[A|Acc],Gs,As):-
		controller_tuple(Q0,O,A,Q1)
		,environment_action(Fs,A,O1,Fs1)
		,executor_bck(Fs1,Q1,O1,Acc,Gs,As).
	\end{minted}
	\caption{Backtracking FSC Executor. $Fs$: initial fluents. $Q0$: current
	controller state label. $O$: current observation label. $A$: Next action
	label. $Acc/As$: Action label accumulator. $Gs$: Goal fluents. $Q1,O1$:
	next state and observation labels.}
	\label{fig:backtracking_executor}
\end{figure}

\begin{figure}[t]
	\centering
	\begin{minted}{prolog}
	executor_rev(Gs,_T1,_Ss,As,Gs,As):- !.
	executor_rev(_Fs,_T1,[],As,_Gs,As):- !.
	% T1 is a partially ground 4-tuple (q,o,A,Q')
	executor_rev(Fs,T1,Ss,[T0|Acc],Gs,As):-
		% All ground instances of T1 are pushed to the stack.
		push_tuples(T0,T1,Ss,Ss_1)
		% T2 is a ground instance of T1 (q,o,a,q')
		,pop_tuple(Ss_1,T2,Ss_2)
		% T3 is a new, partially ground 4-tuple (q',o',A',Q'')
		,environment_state_transition(Fs,T2,Fs1,T3)
		,push_reverse(T2,Ss_2,Ss_r)
		,executor_rev(Fs1,T3,Ss_r,[T2,T0|Acc],Gs,As).
	\end{minted}
	\caption{Reversing FSC Executor. $Fs$: initial fluents. $T1$: current
	FSC 4-tuple, partly ground. $Ss$: stack of FSC 4-tuples. $T0$: last FSC
	tuple, fully ground. $Acc/As$: FSC 4-tuple accumulator. $Gs$: goal
	fluents. \emph{In body literals}: $T2$: next fully ground instance of
	$T1$. $T3$: new, partly ground FSC 4-tuple.
	}\label{fig:reversing_executor}
\end{figure}

\subsubsection{Backtracking Executor}
\label{Backtracking Executor}

We implement a Backtracking Executor that performs the loop described in Section
\ref{Executors} in the simplest possible way, as a recursive procedure in
Prolog, that is allowed to backtrack and obtain further pairs $(a,q')$ while the
goal state has not been reached. Thus, the Backtracking Executor implicitly uses
Depth First Search with backtracking (DFS) and a stack managed by the Prolog
runtime, to "jump back" to locations on a map where alternative actions can be
taken. The Backtracking Executor is listed, as Prolog code, in Figure
\ref{fig:backtracking_executor}.

\subsubsection{Reversing Executor}
\label{Reversing Executor}

Backtracking causes an environment to be "reset" to a previous state, e.g. with
respect to an agent's coordinates in the environment. That may not be possible
in all environments. In particular, simulator-like virtual environments such as
the Basic Environment may allow backtracking but real-world environments such as
the physical world, do not. Thus the Backtracking Executor cannot be used in the
real world.

To allow FSCs to be used in real-world environments we implement a Reversing
Executor that manages a stack explicitly, pushing on the stack the reverse of
each FSC 4-tuple executed. When no more tuples can be pushed on the stack the
first 4-tuple popped from the stack is the reverse of the last executed 4-tuple,
causing the FSC to explicitly retrace its steps. When the last reverse-tuple is
popped from the stack, the next 4-tuple is a not-yet tried alternative and the
agent explores a new path, or the stack is empty and execution terminates.

The reverse of a tuple is determined by an FSC. Only the reverse of a pair
$(a,q')$ need be defined, rather than the reverse of every tuple\footnote{This
does not violate the model-free nature of an FSC: the reverse-pair relation is
specific to an FSC and does not depend on the environment. The same relation can
be used to avoid oscillations, e.g. to avoid following a "down" with an "up"
action.}. Not all action-state pairs will have a reverse, thus not all decisions
can be reversed.

The Reversing Executor is listed, as Prolog code, in Figure
\ref{fig:reversing_executor}.

\subsubsection{Grid SLAM}
\label{Grid SLAM}

\begin{figure}[t]
	\centering
	\begin{tabular}{ll}
		\subfloat[Maze A \label{fig:maze_a_slam_path}]{\includegraphics[width=0.20\columnwidth]{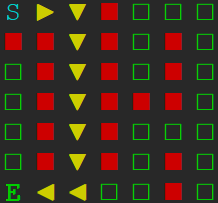}} &
		\subfloat[Maze A SLAM \label{fig:maze_a_slam_map}]{\includegraphics[width=0.23\columnwidth, height=30mm, keepaspectratio]{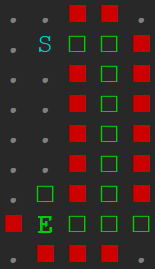}} \\
	\end{tabular}
		\quad
	\begin{tabular}{ll}
		\subfloat[Maze B \label{fig:maze_b_slam_path}]{\includegraphics[width=0.20\columnwidth]{maze_b_slam_path.png}} &
		\subfloat[Maze B SLAM \label{fig:maze_b_slam_map}]{\includegraphics[width=0.23\columnwidth]{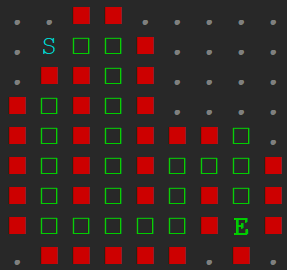}} \\
	\end{tabular}
	\caption{Grid SLAM: mapping of an unseen environment during FSC
	execution with Reversing Executor (yellow arrows). Grey dots:
	unobservable tiles.}\label{fig:slamming}
\end{figure}

When a map has wide areas of passable tiles an FSC can be stuck going around in
loops. To avoid this we implement Simultaneous Localisation and Mapping (SLAM)
as a Grid Master predicate used by an executor to update a SLAM-constructed map
by placing obsevation labels on a grid that is continuously expanded as the FSC
explores an environment. A SLAM executor marks the last location of an FSC on
the SLAMming map as visited and an FSC is only allowed to re-visit a location
when reversing course, with a Reversing (SLAM) executor, thus avoiding cycles.
Details of the SLAM implementation are left for future work for lack of space.
Figure \ref{fig:slamming} illustrates SLAMming maps constructed with our method.

\section{Experiments}
\label{Experiments}

We show empirically that a grid navigation solver can be learned by MIL and used
to learn an FSC capable of solving the same problems as the solver.

We carry out experiments in two types of environments with grid-based maps. The
first type of environment are maps of mazes generated by the Grid Master
\emph{MaGe} module. The second type of environment, illustrated in Figure
\ref{fig:solver_vs_controller} are maps of open areas with obstacles, modelling
lakes with islands, or caves, generated by the dungeon map generator in the
R-language package \emph{r.oguelike} \cite{r.oguelike}. We refer to the latter
type of environment as "Lake maps". 

Each MaGe map has a start and end tile placed automatically, at random. For Lake
Maps we find it difficult to generate new maps programmatically, so we
pre-generate 10 maps of dimensions $20 \times 20$ and create new planning
problems by randomly placing a start and end tile in each map for each new
problem. 

Both types of environment are accessed via the Grid Master Basic Environment as
described in Section \ref{Grid Master}. The salient difference between the two
types of environment is that MaGe maps have no open areas of passable tiles
where an agent can get stuck in a loop ("plazas"), whereas Lake Maps are
predominantly made of such plazas. The MaGe maps thus effectively test the
ability of an agent to find a path through a fully-connected graph without
cycles, wheras the Lake maps test the ability of an agent to find a path through
a graph with cycles.

\subsection{Learning a Model-Based Solver with MIL}
\label{Learning a Model-Based Solver with MIL}

\begin{figure}[t]
	\centering
	\begin{tabular}{l}
		\subfloat[Zero map\label{fig:maze_zero}]{\includegraphics[width=0.10\columnwidth]{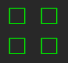}} \\
	\end{tabular}
	\subfloat[Zero map MIL Planning Problem\label{tab:solver_actions_maze_zero}]{
		\begin{tabular}{l}
			\textbf{Second-Order Definite Clauses $M$:} \\
			$(Identity)\; P(x,y) \leftarrow Q(x,y)$ \\
			$(Tailrec)\; P(x,y) \leftarrow Q(x,z), P(z,y)$ \\
			\textbf{Actions $\mathcal{A}$:} \\
			$Step\_down([zero,0/1,f],[zero,0/0,f])$ \\
			$Step\_down([zero,1/1,f],[zero,1/0,f])$ \\
			$Step\_left([zero,1/0,f],[zero,0/0,f])$ \\
			$Step\_left([zero,1/1,f],[zero,0/1,f])$ \\
			$Step\_right([zero,0/0,f],[zero,1/0,f])$ \\
			$Step\_right([zero,0/1,f],[zero,1/1,f])$ \\
			$Step\_up([zero,0/0,f],[zero,0/1,f])$ \\
			$Step\_up([zero,1/0,f],[zero,1/1,f])$ \\
			\textbf{Example $e$:} \\
			$s([zero,x_s/y_s,t_s],[zero,x_e/y_e,t_e).$ \\
		\end{tabular}
	}
	\caption{MIL Planning Problem for Map Zero.} \label{fig:zero_map_planning_problem}
\end{figure}

We find that we can learn a general grid navigation solver with MIL from a
single example of a simplified grid-based map consisting of only four passable
(floor) tiles. We name this map \emph{Zero} and illustrate it in Figure
\ref{fig:maze_zero}.

The initial, uninstantiated model for the target solver is listed in Figure
\ref{tab:grid_solver_model}. To instantiate this model we use Grid Master's
Action Generator module to generate the set of all 8 possible ground actions in
the Zero map. We manually create a \emph{generalised} training example where
each fluent is left as a non-ground term, except for the map identifier. The
resulting MIL Planning Model is listed in Figure
\ref{tab:solver_actions_maze_zero}. 

We train Louise on the MIL Planning Problem listed in Figure
\ref{tab:solver_actions_maze_zero} and obtain the solver listed in Figure
\ref{tab:grid_solver}. Inspecting the logic program in Figure
\ref{tab:grid_solver} should leave no doubt that it is a general procedure for
grid map navigation which systematically tries every available direction,
recursively, until it reaches a goal state defined by the second argument in
each head literal. Nevertheless we test this learned solver on two sets of
environments: 100 maze maps with dimensions 100 $\times$ 100; and 10 $\times$ 50
Lake maps (i.e. we generate 50 instances for each of the 10 Lake
maps)\footnote{We test fewer Lake map problems because the solver is executed in
SWI-Prolog with tabling which avoids looping in plazas but requires a lot of RAM
for experiments.}. Results are listed in the \emph{Experiment 1} rows of Table
\ref{tab:experiment_results}.

\begin{table}[t]
	\centering
	\begin{tabular}{lllllll}
		\textbf{Experiment} & \textbf{Agent} & \textbf{Environment} & \textbf{Dimensions} & \textbf{Instances} & \textbf{Solved} & \textbf{Steps} \\
		\midrule
		Experiment 1 & Solver & MaGe map & 100 $\times$ 100 & 100 & 100\% & 816.04 \\ 
		             & & Lake map & 20 $\times$ 20 & 10 $\times$ 50 & 100\% & 11.23 \\ 
		\midrule
		Experiment 2 & FSC-BT & MaGe map & 100 $\times$ 100 & 100 & 100\% & 816.04 \\ 
		             & FSC-RE & MaGe map & 100 $\times$ 100 & 100 & 100\% & 3942.78 \\ 
			     & FSC-BT(S) & Lake map & 20 $\times$ 20 & 10 $\times$ 50 & 92\% & 69.54 \\ 
			     & FSC-RE(S) & Lake map & 20 $\times$ 20 & 10 $\times$ 50 & 100\% & 132.98 \\ 
		\bottomrule
	\end{tabular}
	\caption{Experiment results. Instances: number of environment instances
	tested. Solved: percentage of instances solved. Steps: mean number of
	actions taken by an agent navigating between start and end tiles.
	FSC-BT: Backtracking executor. FSC-RE: Reversing executor. FSC-BT(S)
	Backtracking SLAM executor. FSC-RE(S) Reversing SLAM executor.}
	\label{tab:experiment_results}
\end{table}

\subsection{Learning a Model-Free Controller with MIL}
\label{Learning a Model-Free Controller with MIL}

\begin{figure}[t]
	\centering
	\begin{tabular}{cccccccc}
		\subfloat[\label{fig:pppp}]{\includegraphics[height=12mm, keepaspectratio]{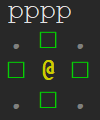}} &
		\subfloat[\label{fig:pppu}]{\includegraphics[height=12mm, keepaspectratio]{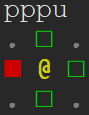}} &
		\subfloat[\label{fig:ppup}]{\includegraphics[height=12mm, keepaspectratio]{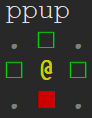}} &
		\subfloat[\label{fig:ppuu}]{\includegraphics[height=12mm, keepaspectratio]{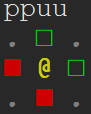}} &
		\subfloat[\label{fig:pupp}]{\includegraphics[height=12mm, keepaspectratio]{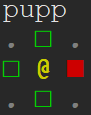}} &
		\subfloat[\label{fig:pupu}]{\includegraphics[height=12mm, keepaspectratio]{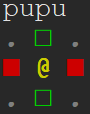}} &
		\subfloat[\label{fig:puup}]{\includegraphics[height=12mm, keepaspectratio]{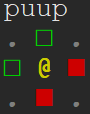}} &
		\subfloat[\label{fig:puuu}]{\includegraphics[height=12mm, keepaspectratio]{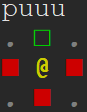}} \\
		\subfloat[\label{fig:uppp}]{\includegraphics[height=12mm, keepaspectratio]{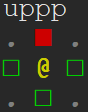}} &
		\subfloat[\label{fig:uppu}]{\includegraphics[height=12mm, keepaspectratio]{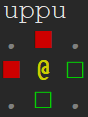}} &
		\subfloat[\label{fig:upup}]{\includegraphics[height=12mm, keepaspectratio]{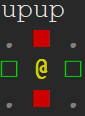}} &
		\subfloat[\label{fig:upuu}]{\includegraphics[height=12mm, keepaspectratio]{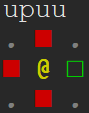}} &
		\subfloat[\label{fig:uupp}]{\includegraphics[height=12mm, keepaspectratio]{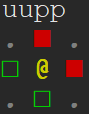}} &
		\subfloat[\label{fig:uupu}]{\includegraphics[height=12mm, keepaspectratio]{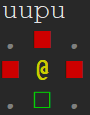}} & 
		\subfloat[\label{fig:uuup}]{\includegraphics[height=12mm, keepaspectratio]{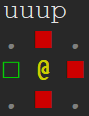}} &
		\\ 
	\end{tabular}
	\caption{Observation Matrices for Grid Navigation FSC. Yellow $@$ sign:
	agent.} \label{fig:observation_matrices}
\end{figure}

The ambiguity illustrated in Figure \ref{fig:maze_ambiguity} suggests a large
number of diverse environments would be needed to use our solver to train a
controller that can solve the same navigation problems as the solver. We observe
that the effect of ambiguity is that not all observation labels will be included
in FSC behaviours generated by the solver and so a complete mapping between
observation and action labels will not be seen in training. Following from this
observation we use the set of observation labels we want the target FSC to
recognise, to generate 15 training maps of dimensions $3 \times 3$, one for each
of the 15 observation labels in the target FSC, excluding the impossible $uuuu$
(i.e. all-unpassable) label. We call these training maps, illustrated in Figure
\ref{fig:observation_matrices}, \emph{Observation Matrices}. 

Solving each Observation Matrix with our learned solver, given the model in
Table \ref{tab:solver_fsc_model}, we generate one to four FSC behaviours, each
mapping an observation label to the up to four action labels that correspond to
it. We illustrate this mapping in supplemental material for lack of space.

We use this set of behaviours, and the controller labels listed in Figure
\ref{tab:controller_model} to compose a MIL-learning problem and learn an FSC
with Controller Freak, as described in Section \ref{Controller Freak}. The
controller learned in this way is a non-deterministic FSC comprising 128
4-tuples in a subset of all 960 4-tuples in $Q \times O \times A \times Q$. We
list this FSC in supplemental material for lack of space.

We test the learned FSC on the same planning problems as we tested the learned
solver, using each of our four executors. Results are listed in the Experiment 2
rows of Table \ref{fig:observation_matrices}.

We make two observations. First, on the MaGe map the Backtracking executor
(FSC-BT) returns paths with the same average number of steps as the solver. The
solver is executed by DFS with backtracking like the Backtracking executor, so
they find the same paths. Essentially the Backtracking executor plans, without a
model. Second, the Backtracking-SLAM executor (FSC-BT(S)) takes a long time to
solve some Lake maps because it keeps trying long, spiralling paths. We impose a
time limit of 300 sec. to this executor only to measure the effect and find that
it times out 8\% of the time. With all other executors, the FSC learned from the
solver, solves the same problems as the solver.

\section{Conclusions and Future Work}
\label{Conclusions and Future Work}

We have shown how to train a model-based solver and use it to train a model-free
controller with MIL. The two agents complement each other in that the solver can
plan ahead but needs advance knowledge of an environment whereas the controller
doesn't need advanced knowledge but can't plan and must explore the environment
instead. The solver is learned from a single, generalised example and the
controller from 15 examples automatically generated from the controller's
observation labels. The controller is a new kind of Nondeterministic Finite
State Controller and can solve the same problems as the solver when executed by
the right kind of executor. A SLAM-ming executor is needed to avoid loops in
environments with open areas. We have focused on solvers and controllers for
navigation on grids and have implemented a pair of new Prolog libraries used to
learn solvers and controllers and execute them on grid maps. 

\subsubsection{Limitations and Future Work}

We have only studied the ability of MIL systems to learn solvers and equivalent
FSCs for grid navigation problems. It remains to be seen whether the approach we
have described can be further applied to environments that are not grids, and
tasks other than navigation. Further, connecting our solver and FSC to a robot's
sensors and actors and solving problems in physical, rather than virtual,
environments remains to be done.

\begin{credits}
\subsubsection{\ackname} The author would like to acknowledge the EPSRC grant
no. EP/X030156/1 from which her post-doctoral research wass funded. The MIL team
at the University of Surrey and Drs. Alan Hunter and Alfie Treloar of the
University of Bath have provided patient, vital feedback during the development
of this work.

\subsubsection{\discintname} No competing interests.  \end{credits}
%
%

\bibliographystyle{splncs04}
\bibliography{mybib}

\end{document}